\documentclass{article}

\usepackage{PRIMEarxiv}

\usepackage[utf8]{inputenc} 
\usepackage{amsmath,amssymb,amsfonts}%
\usepackage[T1]{fontenc}    
\usepackage{hyperref}       
\usepackage{url}            
\usepackage{booktabs}       
\usepackage{amsfonts}       
\usepackage{nicefrac}       
\usepackage{microtype}      
\usepackage{lipsum}
\usepackage{fancyhdr}       
\usepackage{graphicx}       
\graphicspath{{media/}}     
\usepackage{amsfonts}
\usepackage{siunitx}
\usepackage{caption}
\usepackage{subcaption}
\usepackage{graphicx}
\usepackage{multirow}
\usepackage{hyperref}
\usepackage{tikz}
\usepackage{soul}
\soulregister\ref7
\soulregister\cite7

\title{Vascular Geometry Characterization for AI–Based Endovascular Navigation
\thanks{\textit{\underline{Citation}}: 
\textbf{Wu, HR., Robertshaw, H., Dwyer-Joyce, L. et al. Vascular geometry characterization for AI-based endovascular navigation. Int J CARS (2026). \url{https://doi.org/10.1007/s11548-026-03742-9}}} 
}

\author{
    Han-Ru Wu\thanks{Equal contribution} \\
    Department of Medical Imaging Science\\
    National Taiwan University Hospital\\
    Taiwan \\
    \And
    Harry Robertshaw\textsuperscript{\(\dag \)}, Thomas C. Booth, Alejandro Granados\thanks{Corresponding author: \texttt{alejandro.granados@kcl.ac.uk}} \\
    Surgical \& Interventional Engineering \\
    School of Biomedical Engineering \& Imaging Sciences \\
    Kings College London \\
    UK\\
   \And
     Lisa Dwyer-Joyce \\
  Chelsea \& Westminster Hospital\\
  UK\\
}

\begin{document}
\maketitle

\begin{abstract}
    \textbf{Purpose:} Mechanical thrombectomy (MT) is a time-critical intervention for acute ischemic stroke; however, access remains limited due to a shortage of neuroradiologists and specialized centers. Reinforcement learning (RL) offers potential to automate endovascular navigation and improve accessibility, yet current models lack standardized frameworks to assess navigation difficulty for model training and evaluation. This study aims to identify vascular metrics associated with navigation difficulty and to develop an automated pipeline for quantitative vascular feature extraction, enabling future complexity grading. \textbf{Methods:} Vascular trees were segmented from computed tomography angiograms from 61 patients, and vascular metrics including aortic arch type, presence of bovine arch, vessel length, tortuosity, take-off angle, number of reverse curves, were measured using a custom pipeline. A Soft Actor-Critic RL algorithm was used for 120\,s autonomous navigation. Outcomes were analyzed using both mixed effects linear and logistic regression. \textbf{Results:} On the left side, the presence of a bovine arch and aortic arch type~II/III increased navigation time by~30.19\,s and~37.92\,s, respectively, while greater tortuosity ($\beta = 118.20$) further prolonged the procedure and reduced success probability. On the right side, type~II/III arches extended procedure time by~45.94\,s, while each additional reverse curve was associated with~3.96\,s longer navigation time and lower probability of success. \textbf{Conclusion:} These findings demonstrate for the first time that MT agent navigation difficulty is strongly influenced by vascular geometry. The proposed automated pipeline enables objective and quantitative characterization of vascular features, providing a foundation for future development of standardized complexity grading and RL model evaluation, without aiming to demonstrate clinically generalizable autonomous navigation. Our code for automated vascular metrics quantification is available at \url{http://github.com/SurgicalDataScienceKCL/AI-VascularGeometryCharacterisation}
\end{abstract}

\section{Introduction}

    Mechanical thrombectomy (MT) is now an established gold-standard treatment for large-vessel occlusions in acute ischemic stroke, the second leading cause of death worldwide~\cite{feigin_world_2025, jovin_thrombectomy_2015}. However, timely access to MT remains uneven, particularly in rural and remote regions~\cite{hagedorn_rural_2025}. As a result, AI-driven strategies, especially reinforcement learning (RL), have been proposed to support and potentially automate endovascular navigation during MT~\cite{robertshaw_artificial_2024}. Despite the growing popularity of RL-based approaches in endovascular navigation, the lack of standardized vascular metric quantification limits the ability to effectively compare models~\cite{moosa_benchmarking_2025}. This limitation is further compounded by the fact that many existing open-source benchmarks rely on simplified vessel geometries. Consequently, the establishment of an automated and standardized vascular geometry quantification pipeline is crucial for advancing the field.

    Reinforcement learning has increasingly been applied to robotic control and autonomous navigation tasks to address the challenge of achieving safe and timely endovascular navigation~\cite{robertshaw_artificial_2024}. RL-assisted robotic navigation has the potential to support less-experienced specialists in peripheral hospitals, enabling more effective delivery of MT~\cite{robertshaw_artificial_2024}. Recent work has demonstrated RL-based autonomous MT navigation with safe and efficient two-device control in cerebral vessels \textit{in silico}~\cite{robertshaw_reinforcement_2025}. However, current open-source benchmarks largely rely on simplified vessel geometries~\cite{karstensen_learning-based_2024}, and publicly available datasets incorporating real patient vasculatures remain scarce, thereby limiting generalizability to real-world clinical scenarios. Even when benchmarking is performed using real vasculature, vascular geometry and complexity are often not quantified, preventing meaningful interpretation of agent performance relative to navigation difficulty~\cite{moosa_benchmarking_2025}. Consequently, comparisons across studies from different institutions remain challenging, slowing overall research progress.

    Additionally, in the context of MT, multiple clinical studies have suggested that vascular geometry influences navigation difficulty and may contribute to prolonged procedural times, with features such as tortuosity playing a significant role~\cite{snelling_unfavorable_2018}. This underscores the need for a standardized vascular geometry evaluation framework for use in RL model development and evaluation. Moreover, although prior studies have explored associations between vascular anatomy and navigation difficulty, the available evidence remains limited, and there is little consensus regarding which specific vascular features are most strongly associated with prolonged reperfusion  ~\cite{bhogal_initial_2025, miki_combined_2021}. Consequently, no standardized framework for vascular metric quantification currently exists to guide model training or to evaluate navigation difficulty in a consistent and reproducible manner, highlighting a critical gap in current RL-assisted MT research.

    This study aims to systematically characterize vascular features that influence AI-based endovascular navigation difficulty during MT. Using vascular morphological and geometrical parameters derived from computed tomography angiography (CTA) scans, we developed an automated pipeline to quantify vascular geometry and analyze its association with navigation outcomes. Our contributions are as follows: 1) we introduce a pipeline for extracting reproducible vascular features from CTA-derived centerlines, allowing performance comparisons of RL models trained in different vascular environments without requiring raw data sharing (thus preserving patient privacy); 2) we perform a detailed analysis identifying which vascular features are predictive of high or low performance, using a state-of-the-art autonomous RL agent to provide a consistent ‘operator’ across a large number of navigation attempts, expanding the currently limited body of evidence on how specific anatomical characteristics influence navigation in MT, rather than aiming to develop a generalized navigation agent, thereby enabling isolation of anatomical effects and expanding the currently limited evidence on how specific anatomical characteristics influence navigation in MT; 3) we perform our analysis on the largest studied dataset of CTA-derived real patient vasculatures used for autonomous endovascular navigation.

\section{Methods}
    
\begin{figure}[htb!]
            \centering
            \includegraphics[width=1\linewidth, trim={0cm 0cm 0cm 0cm}, clip]{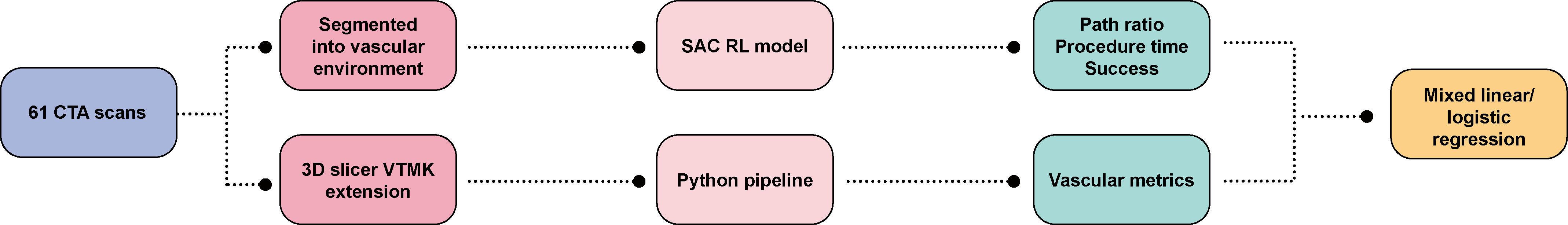}
            \vspace{5pt}
            \caption{Study design. CTA scans are segmented to generate patient-specific vascular environments and centerlines. These are used to (i) train and evaluate an SAC-based endovascular navigation model, yielding navigation outcomes (path ratio, procedure time, and success), and (ii) extract quantitative vascular metrics via an automated Python pipeline. Associations between vascular metrics and navigation outcomes are subsequently analyzed using mixed linear and logistic regression models.}
            \label{fig:study_design}
        \end{figure}

    Our methodology consists of four main steps, as illustrated in Fig.~\ref{fig:study_design}. First, patient vasculature was segmented from CTA scans. Second, vascular morphological and geometrical metrics were automatically quantified using a custom Python-based pipeline. Third, autonomous endovascular navigation was simulated using Soft Actor-Critic reinforcement learning agents. Finally, mixed linear regression models were employed to analyze associations between vascular features and navigation performance outcomes.

    \subsection{Vasculature Segmentation and Centerline Extraction}

        Segmentation of the cerebral vasculature was performed on CTA scans (obtained with UK Research Ethics Committee 24/LO/0057) using \textit{3D Slicer} (v5.6.2) with the \textit{Vascular Modeling Toolkit}. After segmentation, vasculature trees from 61~patients were obtained, including the centerlines, branch endpoints, and vessel radii for each tree.

    \subsection{Quantitative Metric Extraction with Automated Pipeline}

        The automated pipeline was designed to take CTA-derived vascular centerlines as input and automatically extract a set of reproducible morphological and geometrical metrics characterizing the underlying vascular anatomy, as listed below.
    
        Based on current literature, eight vascular morphological and geometrical metrics were included in our study~\cite{snelling_unfavorable_2018, lahlouh_automated_2023}. These were 1) aortic arch type, 2) presence of a bovine arch, 3) length from the aortic origin to the bifurcation of the common carotid artery into internal and external branches, 4) corresponding tortuosity, 5) radii of the brachiocephalic artery (BCA), 6) radii of the right common carotid artery (RCCA), 7) radii of the left common carotid artery (LCCA), and 8) number of reverse curves~\cite{snelling_unfavorable_2018, lahlouh_automated_2023, bhogal_initial_2025, miki_combined_2021}. All metrics were automatically quantified using our custom pipeline (Python, v3.12.4).

\begin{figure}[htbp]
        \centering
        \begin{subfigure}[b]{0.32\linewidth}
            \centering
            \includegraphics[width=\linewidth, trim={0cm 0cm 0cm 0cm}, clip]{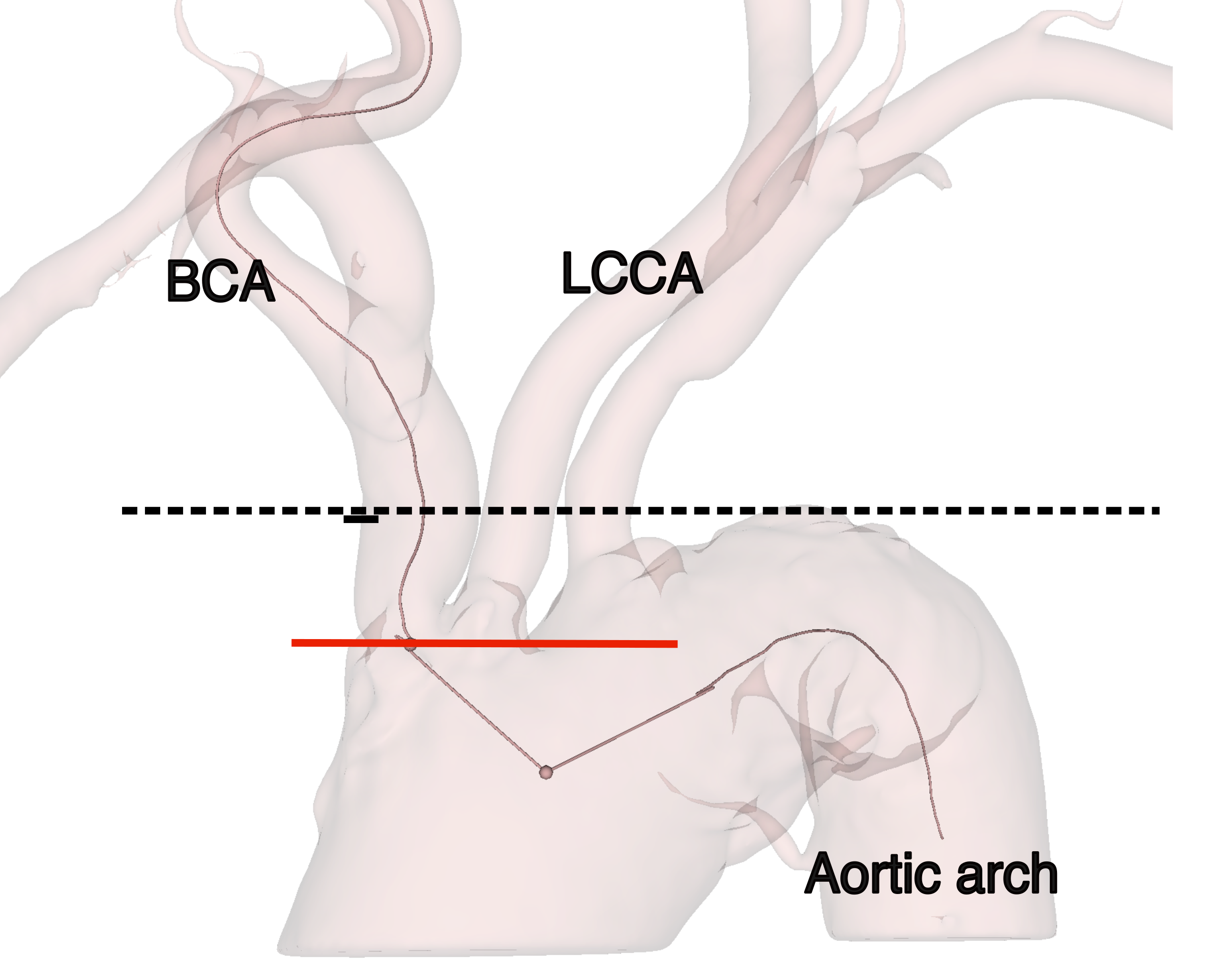}
            \caption{Aortic Arch Type}
            \label{fig:Arch_type}
        \end{subfigure}
        \hfill
        \begin{subfigure}[b]{0.32\linewidth}
            \centering
            \includegraphics[width=\linewidth, trim={0cm 0cm 0cm 0cm}, clip]{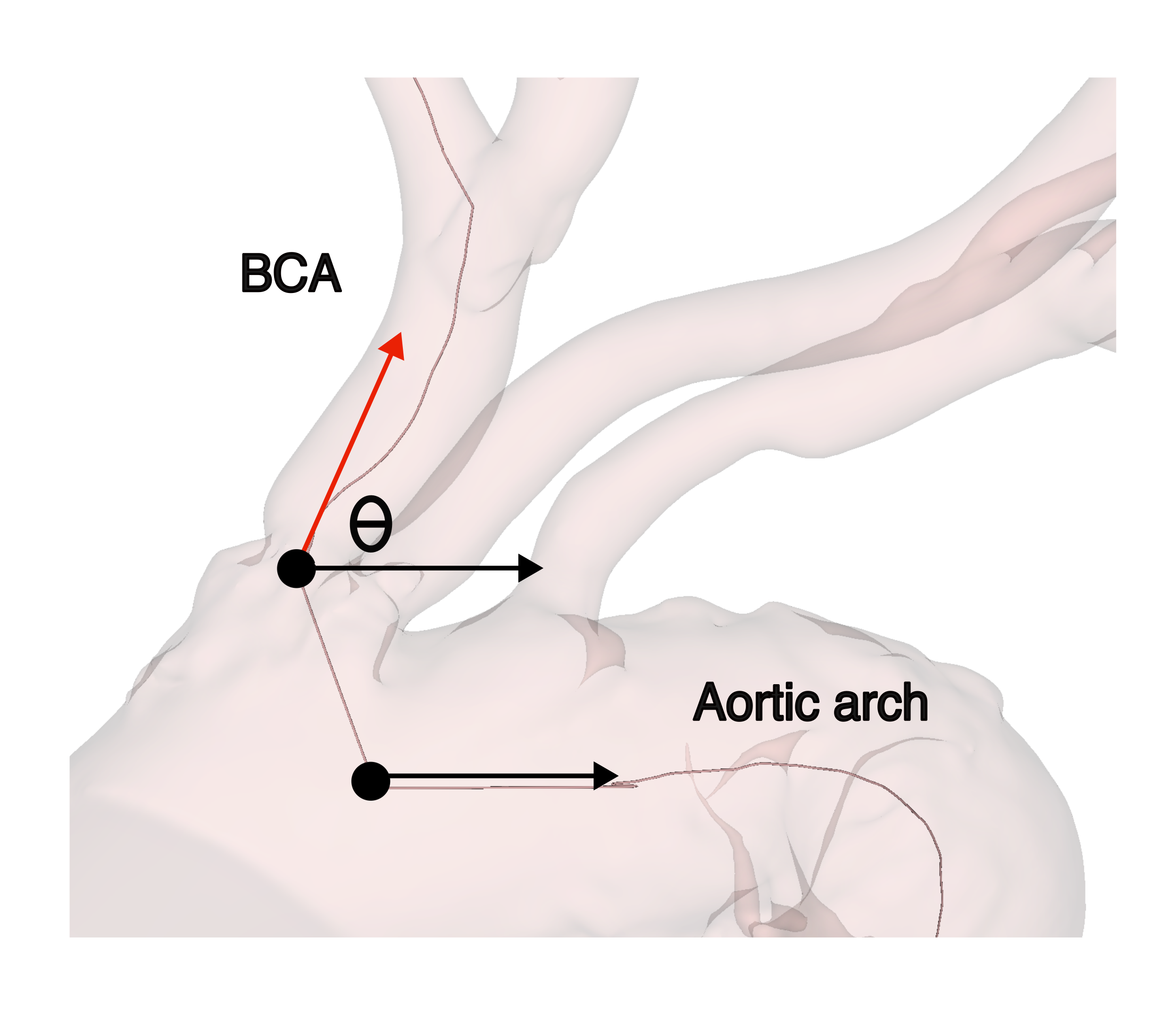}
            \caption{BCA Take-off Angle}
            \label{fig:BCA}
        \end{subfigure}
        \hfill
        \begin{subfigure}[b]{0.24\linewidth}
            \centering
            \includegraphics[width=\linewidth, trim={0cm 0cm 0cm 0cm}, clip]{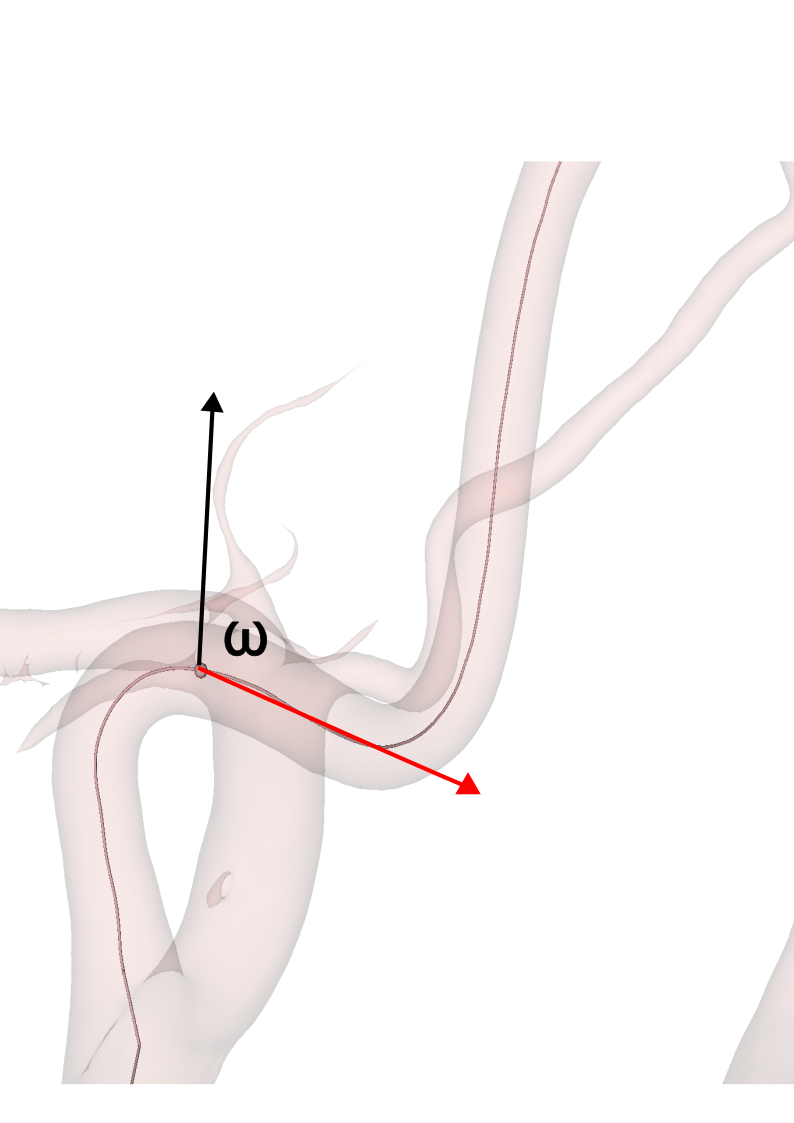}
            \caption{Reverse Curve}
            \label{fig:Reverse}
        \end{subfigure}
    
        \caption{Vascular geometric features: (a) Aortic arch type, defined based on the vertical distance between the highest point of the aortic arch (black dashed line) and the level of the brachiocephalic artery (BCA) origin (red line); (b) BCA take-off angle ($\theta$), computed as the angle between the vector along the BCA (red arrow) and the vector along the aortic arch (black arrow); and (c) Reverse curve angle ($\omega$), computed as the angle between the global vasculature direction vector obtained via singular value decomposition (SVD) of the entire vascular tree (black arrow) and the local vessel direction vector (red arrow).  A reverse curve is defined when $\omega > 100^\circ$.}
        \label{fig:Aortic_features}
    \end{figure}

    \noindent\textbf{Aortic Arch Type.}  
        Aortic arch type was classified into three categories based on the relative position of the BCA origin and the highest point of the arch, and the mean diameter of the LCCA (Fig.~\ref{fig:Arch_type}). Aortic arch type was defined by the vertical distance between the BCA origin and the arch apex: $<1$ LCCA diameter (Type~I), $1$--$2$ LCCA diameters (Type~II), and $>2$ LCCA diameters (Type~III).
  
        The overall direction of the aortic centerline was first determined using singular value decomposition. A plane perpendicular to this direction was then identified, and the point on the aortic arch with the greatest distance from this plane was identified as the highest point of the arch. The mean LCCA diameter was calculated from the vessel segment located $10-20$~mm distal to its origin. 
        
    \noindent\textbf{Bovine Arch.}
        The presence of a bovine arch was determined by assessing whether the BCA and LCCA shared a common origin. This metric was computed by evaluating the geometric intersection between the BCA and LCCA centerlines. In cases where a common point was detected, the vasculature was classified as exhibiting a bovine arch pattern; otherwise, it was classified as a normal aortic configuration. 
    
    \noindent\textbf{Length and Tortuosity.} 
        Length on the right side was measured from the origin of the BCA to the point where the CCA bifurcates into the ICA and ECA. On the left side, it was measured from the origin of the LCCA to its bifurcation point in non-bovine arches, and from the common origin with the BCA in bovine arches. Tortuosity was calculated as the ratio $(d/L)$ of geodesic length, $d$ to Euclidean length.

    \noindent\textbf{Radii.}
        The radius at each arterial origin (BCA, RCCA, and LCCA) was defined as the mean radius.

    \noindent\textbf{Take-off Angle.}
        The take-off angle, representing the branching orientation of the supra-aortic vessels, was defined as the angle between two vectors: one extending 10 mm distally from the arterial origin (BCA for the right side and LCCA for the left side) along its centerline (Fig.~\ref{fig:BCA}), and the other aligned with the direction of the aortic arch centreline.
    
    \noindent\textbf{Reverse Curves.}
        Each segment of the vessel centerline was compared with the vertical axis of the entire vascular tree; segments forming an angle greater than $100^{\circ}$ were classified as reverse curves, indicating that the guidewire must initially advance in the opposite direction before proceeding toward the target vessel (Fig.~\ref{fig:Reverse}). The angle was computed at 5~mm intervals along the vessel centerline to capture local directional changes.

    \subsection{Autonomous Navigation Agents}
    
        \subsubsection{Soft Actor-Critic}
    
            Autonomous navigation agents were trained to provide a consistent ‘operator’ across the large number of evaluation episodes required for an effective regression model to be built. To isolate the effect of vascular geometry on navigation performance, a separate RL agent was trained for each patient-specific vascular model under identical training conditions. This design avoids confounding performance differences arising from cross-anatomy policy generalisation and allows navigation outcomes to be evaluated within each anatomical environment independently. All models in this study were trained using a Soft Actor-Critic (SAC) controller adapted from previous work (Available at \url{https://github.com/lkarstensen/stEVE})~\cite{karstensen_learning-based_2024}.The architecture includes a Long Short-Term Memory layer for learning trajectory-dependent state representations, which has been shown to allow probing of the correct vessel when the target branch is not unambiguously located from the target coordinates~\cite{karstensen_recurrent_2023}. The architecture also includes feedforward layers for controlling the devices. The controller takes observations as input, and a Gaussian policy network outputs mean ($\mu$) and standard deviation ($\sigma$) for expected actions, representing the catheter's and guidewire's rotation and translation. During training, actions are sampled from the $\sigma$, but for evaluation, $\mu$ is used directly for deterministic behavior. 

            A dense reward function was used during training as shown in Equation~\ref{eq:R}~\cite{robertshaw_autonomous_2024}. \textit{Pathlength} is defined as the distance between the guidewire tip and the target along the centerlines of the arteries, with $\Delta\text{pathlength}$ representing the change in pathlength at time $t$ from the previous step at time $t=-1$.
    
            \begin{equation}
                R = -0.00015 - 0.001\cdot\Delta\text{pathlength}+\begin{cases}1 & \text{if target reached} \\0 & \text{else} \end{cases}
                \label{eq:R}
            \end{equation}

        \subsubsection{Navigation Task}
    
\begin{figure}[]
                \centering
                \includegraphics[width=0.55\linewidth]{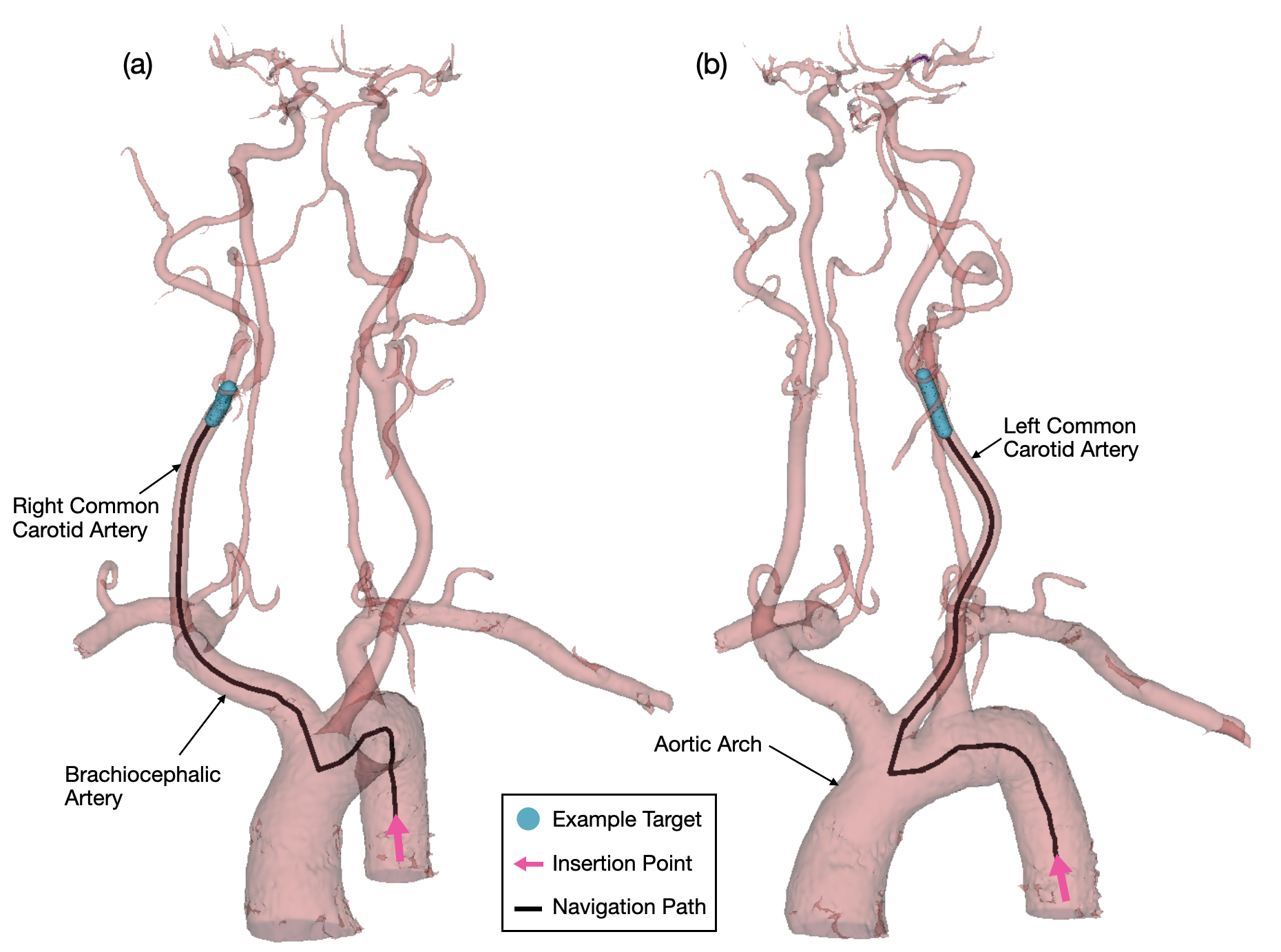}
                \caption{ Autonomous navigation tasks. (a) top of descending aorta to right common carotid artery, and (b) top of descending aorta to left common carotid artery.}
                \label{fig:nav_task}
            \end{figure}

            The two navigation tasks studied are shown in Fig.~\ref{fig:nav_task}. They comprise navigating a guidewire and guide catheter from 1) the top of the descending aorta to the RCCA, and 2) from the top of the descending aorta to LCCA. These are key steps in MT~\cite{crossley_validation_2019}, and previous research has found these tasks to be some of most difficult to navigate autonomously during the intervention~\cite{gee_world_2026}.

        \subsubsection{Training and Evaluation Procedure}
    
            All models were trained under the same conditions. In total, 122 agents were trained, one for every environment (61) and task (2). Agents were trained for up to $1 \times 10^6$ exploration steps, or until they reached 100\% success rate during periodic evaluation (conducted every $5 \times 10^4$ exploration steps) - whichever came first. Training was performed on an NVIDIA DGX A100 node (Santa Clara, California, USA) with 8 GPUs, and took approximately 48 hours per model.

            Each agent was evaluated exclusively on its corresponding vascular environment. For each agent, 20 evaluation episodes were conducted using the same seeds but distinct target locations defined on the same vasculature, with a timeout of 120 s. Performance metrics included success rate, procedure time, and path ratio. Success rate was defined as the proportion of evaluation episodes in which the target was reached. Path ratio quantified progress toward the target and was calculated as the distance traveled by the guidewire tip toward the target divided by the initial distance from the guidewire tip to the target. Procedure time was defined as the elapsed time from the start of navigation to target arrival for successful episodes, or 120 s for unsuccessful episodes.

    \subsection{Mixed-Effects Linear Regression Model and Mixed-Effects Logistic Regression model}
    
        The association between vascular features and certain navigation outcomes such as path ratio and procedure time was analyzed using a mixed-effects linear regression model,  with statistical significance defined as \textit{p}~$<0.05$. Two navigation tasks were evaluated per patient vasculature (one targeting the left side and one the right), each comprising 20 evaluation episodes, resulting in 40 evaluation episodes per patient vasculature. Across 61 patient vasculatures, this yielded a total of 2 × 20 × 61=2440 evaluation episodes. The model incorporated random effects to account for variability between vascular trees while estimating the fixed effects of vascular features on navigation (Equation~\ref{eq:Yi}) success~\cite{bates_fitting_2015}. 

        \begin{equation}
            Y_i = (\beta_0 \mid \text{factor differs among groups}) + \beta_1 x_{1i} + \beta_2 x_{2i} + \dots + \beta_n x_{ni} + \varepsilon_i
            \label{eq:Yi}
        \end{equation}

        In Equation~\ref{eq:Yi}, $\beta_0$ represents the vascular tree--specific intercept, capturing baseline differences in navigation difficulty across vascular trees. The coefficients $\beta_1, \ldots, \beta_n$ denote fixed effects, which are assumed to be constant across vascular trees. The variables $x_{1i}, x_{2i}, x_{3i}, \ldots$ correspond to quantified vascular metrics (e.g., vessel length and tortuosity) for vasculature $i$. The term $\varepsilon_i$ represents the residual error.

        Candidate predictor sets were defined a priori based on anatomical relevance and prior literature. To avoid overfitting given the sample size, the number of predictors per model was restricted, and highly collinear variables were not included in the same model. Different combinations of vascular metrics were evaluated using the Akaike Information Criterion (AIC), which measures the relative quality of statistical models by balancing model fit and complexity~\cite{kimura_minimization_2016}. A lower AIC value indicates a model that better explains the data with fewer parameters, thereby minimizing the risk of overfitting. The model with the lowest AIC was selected for further investigation of the relationship between vascular metrics and navigation outcomes. 

        To capture navigation success as a bounded, binary outcome derived from repeated episodes, a mixed-effects logistic regression model was additionally employed at the episode level~\cite{hu_comparison_1998}. Success was encoded as a binary variable (0/1), and the model was fit using a logit link function, with a random intercept for each vascular tree to account for within-subject correlation. This approach explicitly accounts for both the bounded nature of the outcome and the repeated-measures structure of the data. For all models, statistical significance was set at $p < 0.05$, and all regression analyses and AIC-based model comparisons were performed in R (v4.4.1).

\section{Results}

    Table~\ref{tab:Table1} presents the quantitative measurements of vascular morphology and geometry. Overall, the length from the aortic origin to the point where the CCA bifurcates into the ICA and ECA was longer on the right side than on the left (163.8~mm vs.~145.9~mm) and exhibited greater variability. The navigation pathway along the right route also tended to be more tortuous (1.31 vs.~1.19). Additionally, reverse curves were predominantly observed along the right-side trajectory and demonstrated substantial variation across vascular trees.
    
    \begin{table}[]
\centering
\caption{Vascular morphological, geometrical and demographic metrics.}
\begin{tabular}{lll}
\toprule
 & \textbf{Mean} & \textbf{SD} \\
\textbf{Length on the right/left side} & $163.80/145.88\ \text(mm)$ & $29.99/15.36\ \text(mm)$ \\
\textbf{Tortuosity on the right/left side} & $1.31/1.19$ & $0.24/0.13$ \\
\textbf{BCA Take-off Angle} & $78.61^{\circ}$ & $19.13^{\circ}$ \\
\textbf{LCCA Take-off Angle} & $61.80^{\circ}$ & $18.56^{\circ}$ \\
\textbf{BCA Radius} & $6.74\ \text(mm)$ & $1.41\ \text(mm)$ \\
\textbf{RCCA Radius} & $3.47\ \text(mm)$ & $0.69\ \text(mm)$ \\
\textbf{LCCA Radius} & $4.02\ \text(mm)$ & $0.75\ \text(mm)$ \\
\textbf{Number of reverse curves on the right/left side} & $3.53/0.09$ & $3.13/0.54$ \\
\textbf{Age} & $66.41\ \text(years)$ & $14.22\ \text(years)$ \\
\\[-6pt]
\midrule
\textbf{Non-bovine arch} & \textbf{Bovine arch} &\ \\
$41/61\ (67.21\%)$ & $20/61\ (32.79\%)$ &\ \\
\textbf{Aortic arch type I} & \textbf{Aortic arch type II/III} &\ \\
$52/61\ (85.25\%)$ & $9/61\ (14.75\%)$ &\ \\
\textbf{Male} & \textbf{Female} &\ \\
$30/61\ (49.18\%)$ & $31/61\ (50.82\%)$ &\ \\
\bottomrule
\end{tabular}
\label{tab:Table1}
\end{table}

\begin{figure}[htbp]
        \centering
        \begin{subfigure}[b]{0.28\linewidth}
            \centering
            \includegraphics[width=\linewidth, trim={0cm 1.9cm 0cm 0.8cm}, clip]{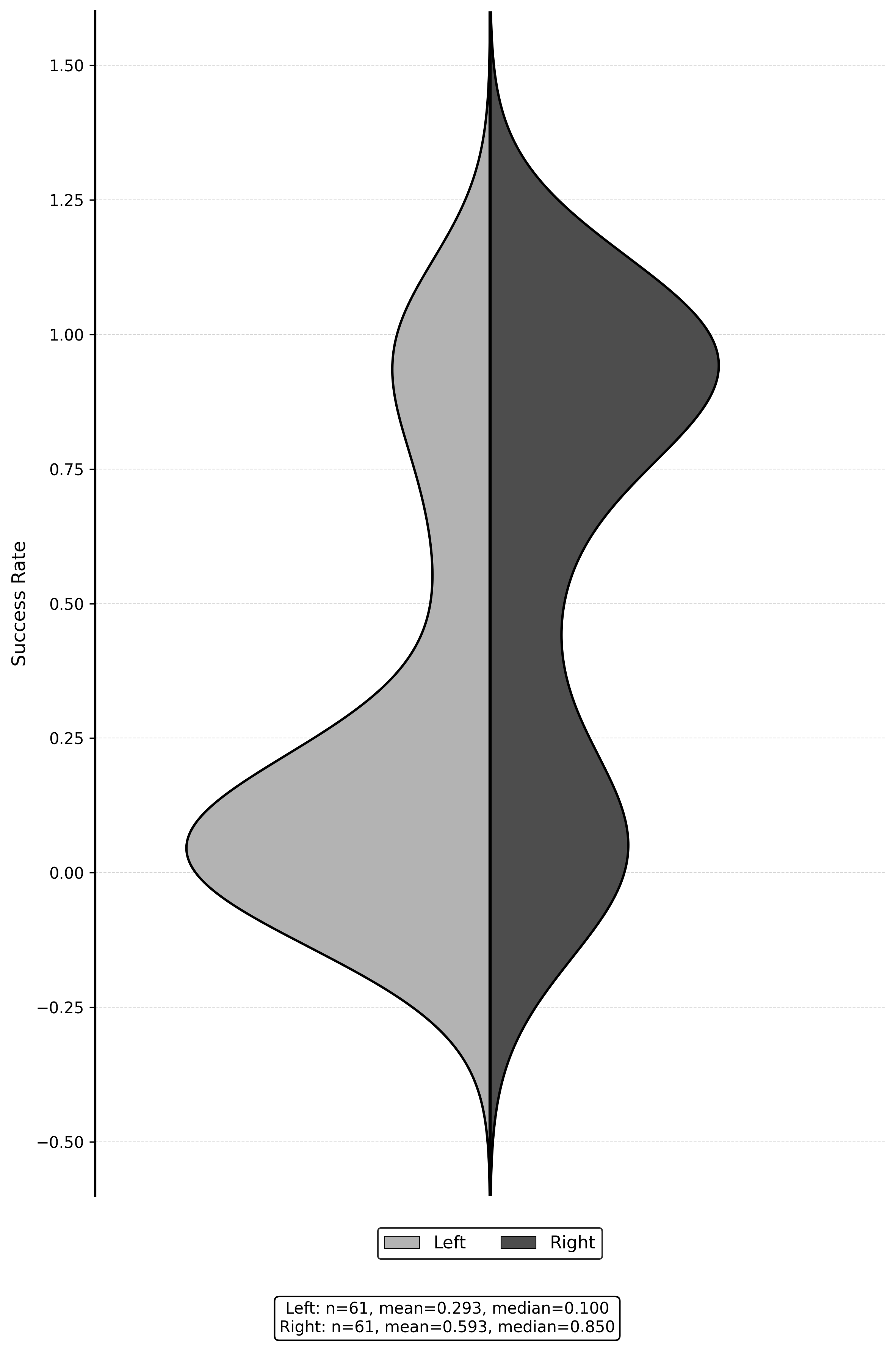}
            \caption{Success rate}
            \label{fig:violin_SR}
        \end{subfigure}
        \hfill
        \begin{subfigure}[b]{0.28\linewidth}
            \centering
            \includegraphics[width=\linewidth, trim={0cm 1.9cm 0cm 0.8cm}, clip]{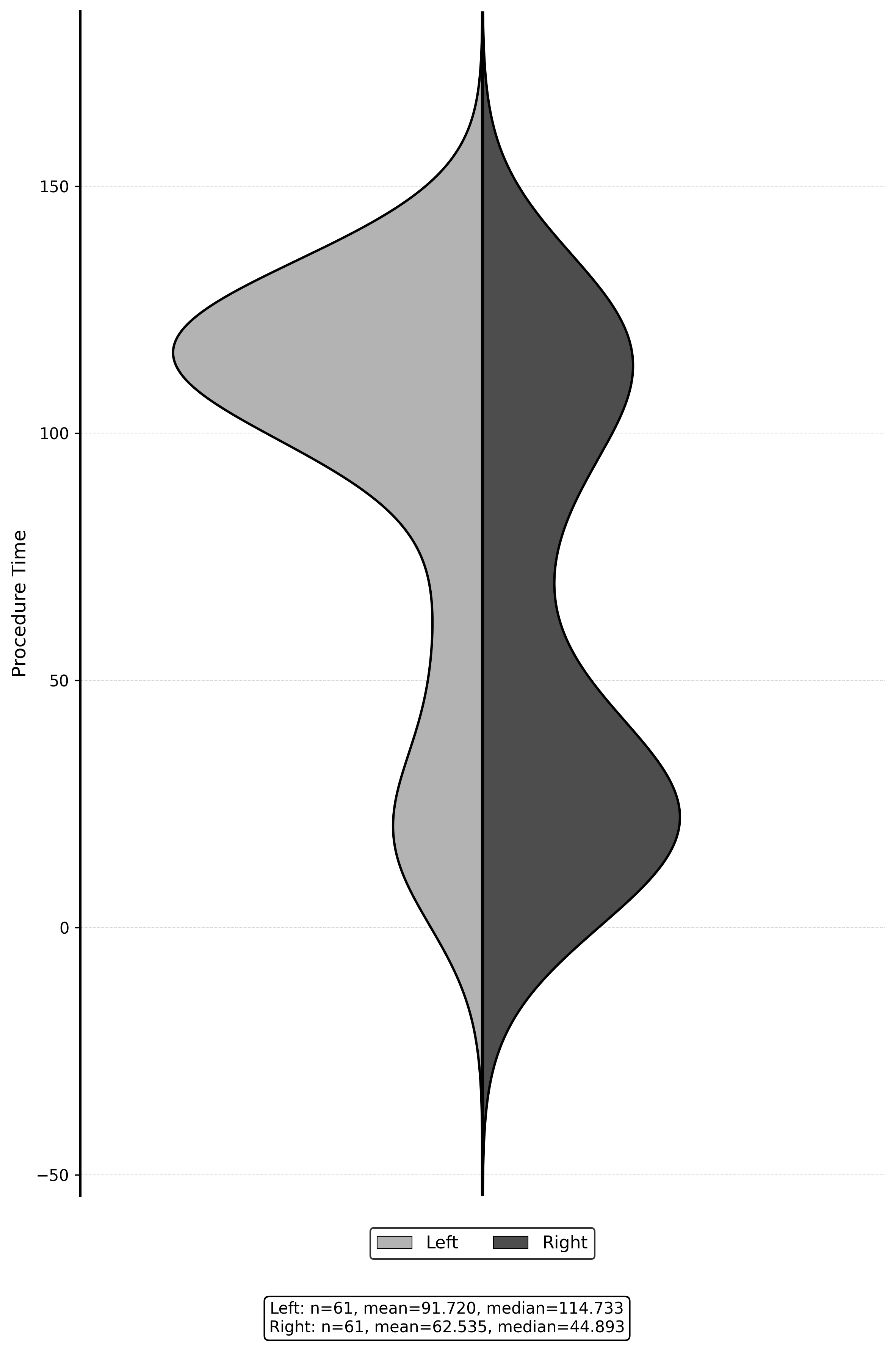}
            \caption{Procedure time}
            \label{fig:violin_PT}
        \end{subfigure}
        \hfill
        \begin{subfigure}[b]{0.28\linewidth}
            \centering
            \includegraphics[width=\linewidth, trim={0cm 1.9cm 0cm 0.8cm}, clip]{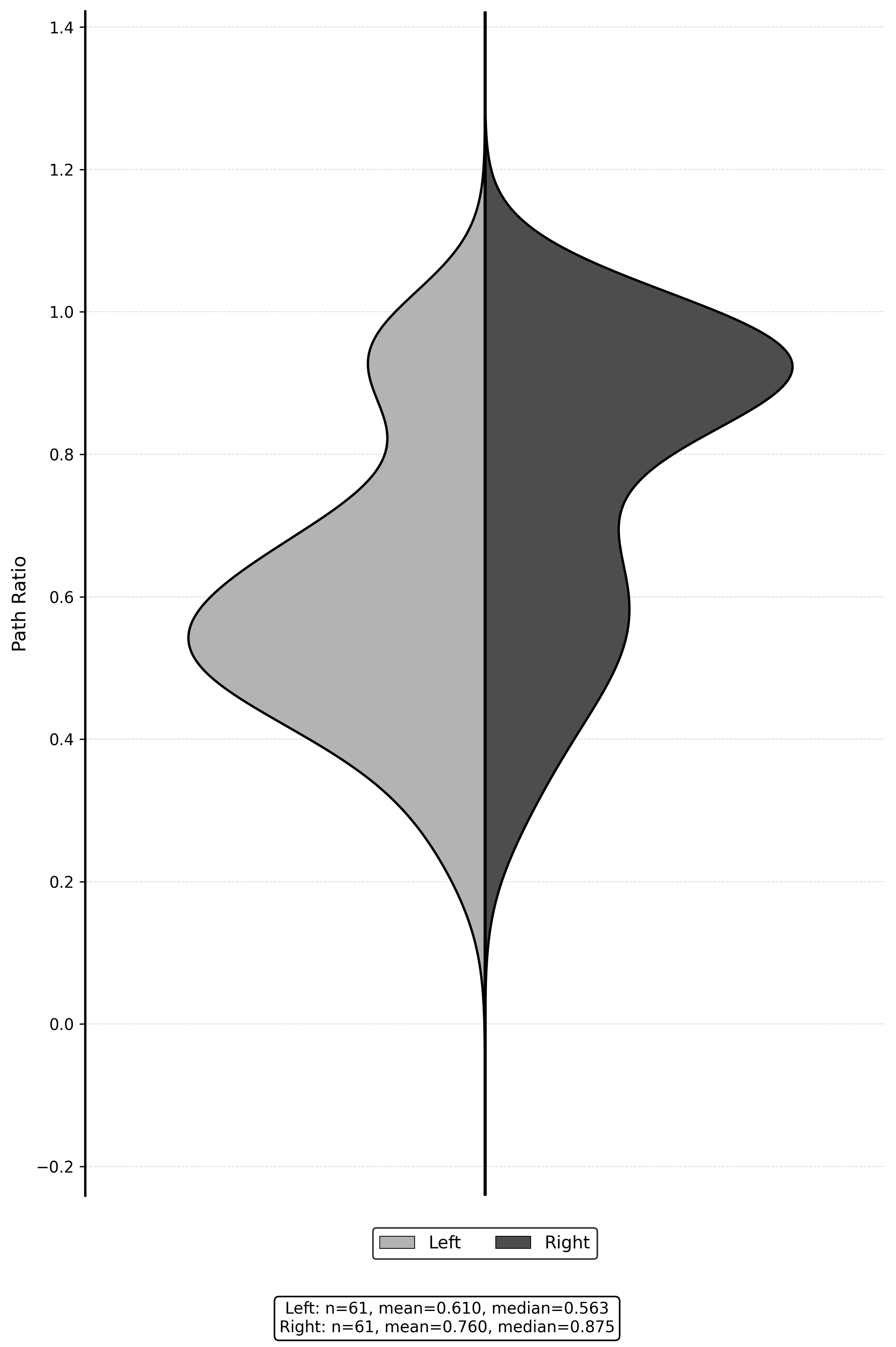}
            \caption{Path ratio}
            \label{fig:violin_PR}
        \end{subfigure}
        \caption{Density distributions for performance metrics on left (light grey) and right (dark grey) side navigation tasks across 61 vasculature.}
        \label{fig:violins}
    \end{figure}

    Plots comparing performance metrics between left and right side tasks across all vasculatures can be seen in Fig.~\ref{fig:violins}. Success rates were higher for right side tasks (mean=0.593, median=0.850) compared to the left (mean=0.293, median=0.100), while procedure time distributions showed that left side tasks needed increased times (mean=91.7\,s, median=114.7\,s) compared to the right (mean=62.5\,s, median=44.9\,s). Path ratio showed that right side tasks finished closer to the target (mean=0.760, median=0.875) when compared to the left side (mean=0.610, median=0.563).

    Table~\ref{tab:Table2} presents the relationship between vascular morphological features and navigation outcomes across tasks. In the left side \textit{path ratio} model, the presence of a bovine arch was significantly associated with a lower path ratio ($\beta = -0.14$, $p < 0.05$), while LCCA radius showed no significant effect ($p = 0.147$). In the left \textit{time} model, longer navigation time was observed in vascular trees with a bovine arch ($\beta = 30.19$, $p < 0.05$), aortic arch type~II/III ($\beta = 37.92$, $p < 0.05$), and increased tortuosity ($\beta = 118.20$, $p < 0.05$). The mixed effect logistic regression model for left-sided \textit{success} yielded consistent results, indicating that vascular configurations associated with prolonged navigation time were also linked to a lower probability of navigation success. Notably, a larger LCCA take-off angle was not significantly associated with procedural time, but was associated with an increased probability of navigation success ($OR = 1.86$, $p < 0.05$).

    For the right side \textit{path ratio} model, the presence of a bovine arch showed a significant negative effect on the path ratio ($\beta = -0.27$, $p < 0.05$), indicating reduced navigation efficiency. In the right \textit{time}  model, arch type and the number of reverse curves were significant predictors of procedure duration. Type~II/III arches were associated with longer navigation times ($\beta = 45.94$, $p < 0.05$), while an increased number of reverse curves ($\beta = 3.96$, $p < 0.05$) similarly prolonged navigation. Regarding the right-sided \textit{success} model, both arch type and the number of reverse curves had significant negative associations with successful navigation ($\beta = -0.45$, $p < 0.05$ and $\beta = -0.04$, $p < 0.05$, respectively), consistent with the findings for procedure time. Similarly to the left side, a larger BCA take-off angle was associated with an increased probability of navigation success($OR = 1.18$, $p < 0.05$).

    \begin{table}[!htbp]
\centering
\footnotesize
\caption{Association between vascular features and outcomes on left (L) and right (R) sides. $^{L}$ and $^{R}$ indicates predictor is significant on L or R, respectively.}
\label{tab:Table2}
\begin{tabular}{lllll}
\toprule
\textbf{Model} & 
\textbf{Predictor} & 
\textbf{\begin{tabular}[c]{@{}c@{}}L: Estimate /\\ Variance (SD)\end{tabular}} & 
\textbf{\begin{tabular}[c]{@{}c@{}}R: Estimate /\\ Variance (SD)\end{tabular}} & 
\textbf{\begin{tabular}[c]{@{}c@{}}P-value \\(L / R)\end{tabular}} \\
\midrule
\multicolumn{5}{l}{\textbf{\textit{Path Ratio}}} \\
Random Effects & Vasculature & 0.035 (0.19) & 0.033 (0.18) & -- \\
               & Residual    & 0.070 (0.26) & 0.060 (0.24) & -- \\
Fixed Effects  & Bovine Arch$^{LR}$ & \textbf{-0.14 (0.05)} & \textbf{-0.27 (0.07)} & $<0.05$ / $<0.05$ \\
               & LCCA Radius & $-0.05$ (0.03) & $-0.02$ (0.01) & 0.147 / 0.06 \\
\midrule
\multicolumn{5}{l}{\textbf{\textit{Time}}} \\
Random Effects & Vasculature & 1189.2 (34.49) & 1189.2 (34.39) & -- \\
               & Residual    & 518.3 (22.77)  & 518.3 (22.77)  & -- \\
Fixed Effects  & Bovine Arch$^{L}$ & \textbf{30.19 (13.2)} & -- & $<0.05$ / -- \\
               & Arch Type$^{LR}$   & \textbf{37.92 (3.12)} & \textbf{45.94 (5.09)} & $<0.05$ / $<0.05$ \\
               & Tortuosity$^{L}$  & \textbf{118.20 (2.94)} & -- & $<0.05$ / -- \\
               & Reverse Curves$^{R}$ & $-22.05$ (2.55) & \textbf{3.96 (1.56)} & 0.06 / $<0.05$ \\
               & BCA Angle   & -- & $-0.46$ (0.27) & -- / 0.09 \\
               & BCA Radius  & -- & $-5.21$ (3.61) & -- / 0.15 \\
\midrule
\multicolumn{5}{l}{\textbf{\textit{Success}}} \\
 Random Effects& Vasculature& 9.18 (3.03)& 12.25 (3.05)&\\
Fixed Effects  & Bovine Arch$^{L}$ & \textbf{-4.16(1.07)}& -- & $<0.05$ / -- \\
               & Arch Type$^{LR}$   & \textbf{-3.95(1.47)}& \textbf{-3.23(1.13)}& $<0.05$ / $<0.05$ \\
               & Reverse Curves$^{R}$ & --& \textbf{-1.63(0.57)}& -- / $<0.05$\\
               & BCA Angle   & -- & \textbf{1.18(0.56)}& -- / $<0.05$\\
 & LCCA Angle & \textbf{1.86(0.53)}& --&$<0.05$ / -- \\
               & BCA Radius  & -- & 0.81(0.55)& --/0.14\\
\bottomrule
\multicolumn{5}{l}{\footnotesize * Success estimates are reported on the log-odds scale.}
\end{tabular}
\end{table}

\section{Discussion}

    This study aimed to evaluate the influence of vascular morphological features on endovascular navigation performance during MT. By analyzing geometric metrics extracted from CTA-derived vasculature, we identified key anatomical characteristics that affect navigation outcomes such as procedure time, and success rate.
    
    Current evidence regarding which vascular features strongly contribute to navigation difficulty remains limited and inconsistent. For instance, parameters such as the BCA and LCCA take-off angles and the radii of arterial branch origins have often been discussed based primarily on expert opinion rather than on systematic, quantitative assessment of objective agent performance~\cite{lahlouh_automated_2023}. Other vascular features such as bovine arch and aortic arch type have not reached consensus ~\cite{snelling_unfavorable_2018, bhogal_initial_2025, miki_combined_2021}. While some studies have suggested that the presence of a bovine arch and aortic arch type II/III are associated with longer procedure times and lower navigation success, others have reported minimal or no significant impact~\cite{snelling_unfavorable_2018,bhogal_initial_2025, miki_combined_2021}. However, in our study, both the right- and left-side routes demonstrated that aortic arch type II/III configurations were significantly associated with prolonged navigation time and reduced success.  
    
    On the left side, the presence of a bovine arch was associated with increased procedural time, whereas no such effect was observed on the right side. This finding is anatomically plausible, as the bovine arch configuration alters the origin of the LCCA, making navigation more complex. In contrast, the RCCA originates independently from the BCA, which explains why right-hand navigation is largely unaffected.

    Our findings on tortuosity partially diverged from prior reports, which linked increased tortuosity to greater navigation difficulty. In our study, this association was observed only on the left side, as right-sided tortuosity was excluded from the final model because its inclusion did not improve model fit based on the AIC criterion. This discrepancy may be explained by differences in study scope and methodology. Our model evaluated navigation only up to the CCA bifurcation, excluding the ICA segment, whereas prior studies assessed tortuosity along the entire vascular pathway from the aortic origin to the ICA~\cite{bilgin_impact_2025, lahlouh_automated_2023}. Notably, studies reporting a significant association between tortuosity and reperfusion outcomes primarily focused on the ICA, where increased tortuosity has been independently linked to longer puncture-to-reperfusion times~\cite{koge_internal_2022}. This discrepancy may also reflect methodological differences, as prior studies relied on manual procedures by human neuroradiologists, whereas our findings are based on reproducible RL-based simulations.

    Regarding reverse curves, evidence linking them to navigation outcomes remains limited. In our study, the right-sided BCA–RCCA pathway exhibited a higher number of reverse curves, which was associated with poorer navigation performance, whereas no such association was observed on the left side. Anatomically, reverse curves may cause the guidewire to deviate from the intended trajectory, requiring transient advancement in the opposite direction before realignment. This side-specific observation may reflect inherent anatomical differences, as reverse curves were less prevalent in left-sided configurations in our dataset.

    This study has several limitations specific to the present work. Navigation outcomes were derived from a RL–based \textit{in silico} navigations rather than human-performed procedures. The factors influencing navigation difficulty in the RL model may not fully align with those perceived by human operators, and the lack of validation against expert human navigation data limits the clinical interpretability of the reported metrics. External validation with real-world clinical data is warranted. Nevertheless, this approach ensures consistency and eliminates variability due to operator experience, while providing a framework for agent validation performing navigation tasks on multiple levels of vascular complexity. Future work should compare RL-derived difficulty metrics against human performance benchmarks, enabling calibration of simulation-based measures to clinically meaningful outcomes. 
    
    Due to inherent constraints of the current RL model, training was performed separately for each vasculature, representing a key limitation of this study, and leaving improved generalization and open-question for future work. While this design ensures an unbiased assessment of navigation difficulty within a given anatomy, it limits insight into real-world deployment scenarios in which a single model must robustly handle diverse vascular anatomies. However, it should be noted that the RL architecture used in this study has previously allowed for \textit{in silico} endovascular navigation on unseen environments~\cite{robertshaw_reinforcement_2025, karstensen_learning-based_2024}. To build on the difficulty features identified in this study, future work could explore whether RL agents augmented with targeted training on high-difficulty anatomies outperform baseline agents in terms of cross-anatomy generalization, providing a potential pathway toward more clinically robust models.
    
    The analysis was restricted to navigation from the aortic arch into the common carotid arteries, which may limit generalizability to later stages of MT (such as the more complex ICA segment). Future work will extend the framework to more distal vascular segments, including the ICA siphon, to improve clinical applicability.
    
    Additionally, although this study was conducted on the largest \textit{in silico} dataset of real patient vasculatures used for autonomous endovascular navigation to date, the dataset size remains modest. This limited statistical power and precluded the use of more data-hungry non-linear modelling approaches, leading to an analysis focused on mixed-effects linear models. As larger datasets become available, future work will explore non-linear relationships between vascular features and navigation performance using more complex machine learning methods.
    
    Finally, clinical variables such as age and sex were not incorporated into the analysis. However, vascular morphology often reflects underlying clinical characteristics. For instance, older individuals tend to exhibit greater vascular tortuosity, which may be indirectly captured in our vascular feature measurements ~\cite{ando_factors_2025}.


\section{Conclusion}

 We demonstrated the feasibility of systematically quantifying vascular geometry and investigated its association with RL-based navigation performance in mechanical thrombectomy. For the first time, we established an automated pipeline that objectively extracts vascular morphological and geometrical features and evaluates their relationship with navigation outcomes. Our findings identify key anatomical factors including aortic arch type, the presence of a bovine arch, and the number of reverse curves that significantly influence navigation difficulty. 
 
 This work highlights the potential of combining automated vascular metric quantification with RL-based simulations to standardize complexity assessment and model evaluation. Importantly, the proposed framework operates on CTA-derived vascular centerlines, enabling performance comparison and benchmarking without sharing raw imaging data, thereby supporting multi-centre collaboration without compromising patient privacy. Here, RL is used as a standardized \textit{in silico} operator to probe anatomy-dependent navigation difficulty, rather than to demonstrate clinically generalizable autonomous navigation.
 
 Future research should expand the dataset, incorporate clinical validation, and extend this framework toward real-world endovascular navigation and large-scale, standardized evaluation of RL models.

\section*{Funding}

Partial financial support was received from the WELLCOME TRUST (Grant Agreement No 203148/A/16/Z), the Engineering and Physical Sciences Research Council Doctoral Training Partnership (Grant Agreement No EP/R513064/1), and the MRC IAA 2021 Kings College London (MR/X502923/1). For the purpose of Open Access, the Author has applied a CC BY public copyright license to any Author Accepted Manuscript version arising from this submission.

\bibliographystyle{unsrt}  
\bibliography{references}

@article{koge_internal_2022,
	title = {Internal {Carotid} {Artery} {Tortuosity}: {Impact} on {Mechanical} {Thrombectomy}},
	volume = {53},
	issn = {0039-2499, 1524-4628},
	shorttitle = {Internal {Carotid} {Artery} {Tortuosity}},
	url = {https://www.ahajournals.org/doi/10.1161/STROKEAHA.121.037904},
	doi = {10.1161/STROKEAHA.121.037904},
	language = {en},
	number = {8},
	urldate = {2026-05-31},
	journal = {Stroke},
	author = {Koge, Junpei and Tanaka, Kanta and Yoshimoto, Takeshi and Shiozawa, Masayuki and Kushi, Yuji and Ohta, Tsuyoshi and Satow, Tetsu and Kataoka, Hiroharu and Ihara, Masafumi and Koga, Masatoshi and Isobe, Noriko and Toyoda, Kazunori},
	month = aug,
	year = {2022},
	pages = {2458--2467},
}

@incollection{gee_world_2026,
	address = {Cham},
	title = {World {Model} for {AI} {Autonomous} {Navigation} in {Mechanical} {Thrombectomy}},
	volume = {15968},
	isbn = {9783032051134 9783032051141},
	url = {https://link.springer.com/10.1007/978-3-032-05114-1_65},
	language = {en},
	urldate = {2026-05-31},
	booktitle = {Medical {Image} {Computing} and {Computer} {Assisted} {Intervention} – {MICCAI} 2025},
	publisher = {Springer Nature Switzerland},
	author = {Robertshaw, Harry and Wu, Han-Ru and Granados, Alejandro and Booth, Thomas C.},
	editor = {Gee, James C. and Alexander, Daniel C. and Hong, Jaesung and Iglesias, Juan Eugenio and Sudre, Carole H. and Venkataraman, Archana and Golland, Polina and Kim, Jong Hyo and Park, Jinah},
	year = {2026},
	doi = {10.1007/978-3-032-05114-1_65},
	pages = {680--690},
}

@article{crossley_validation_2019,
	title = {Validation studies of virtual reality simulation performance metrics for mechanical thrombectomy in ischemic stroke},
	volume = {11},
	issn = {1759-8478, 1759-8486},
	url = {https://jnis.bmj.com/lookup/doi/10.1136/neurintsurg-2018-014510},
	doi = {10.1136/neurintsurg-2018-014510},
	language = {en},
	number = {8},
	urldate = {2026-05-31},
	journal = {Journal of NeuroInterventional Surgery},
	author = {Crossley, Robert and Liebig, Thomas and Holtmannspoetter, Markus and Lindkvist, Johan and Henn, Pat and Lonn, Lars and Gallagher, Anthony Gerald},
	month = aug,
	year = {2019},
	pages = {775--780},
}

@article{karstensen_recurrent_2023,
	title = {Recurrent neural networks for generalization towards the vessel geometry in autonomous endovascular guidewire navigation in the aortic arch},
	volume = {18},
	issn = {1861-6429},
	url = {https://link.springer.com/10.1007/s11548-023-02938-7},
	doi = {10.1007/s11548-023-02938-7},
	language = {en},
	number = {9},
	urldate = {2026-05-31},
	journal = {International Journal of Computer Assisted Radiology and Surgery},
	author = {Karstensen, Lennart and Ritter, Jacqueline and Hatzl, Johannes and Ernst, Floris and Langejürgen, Jens and Uhl, Christian and Mathis-Ullrich, Franziska},
	month = may,
	year = {2023},
	pages = {1735--1744},
}

@misc{karstensen_learning-based_2024,
	title = {Learning-{Based} {Autonomous} {Navigation}, {Benchmark} {Environments} and {Simulation} {Framework} for {Endovascular} {Interventions}},
	copyright = {arXiv.org perpetual, non-exclusive license},
	url = {https://arxiv.org/abs/2410.01956},
	doi = {10.48550/ARXIV.2410.01956},
	urldate = {2026-05-31},
	publisher = {arXiv},
	author = {Karstensen, Lennart and Robertshaw, Harry and Hatzl, Johannes and Jackson, Benjamin and Langejürgen, Jens and Breininger, Katharina and Uhl, Christian and Sadati, S. M. Hadi and Booth, Thomas and Bergeles, Christos and Mathis-Ullrich, Franziska},
	year = {2024},
}

@article{robertshaw_artificial_2024,
	title = {Artificial {Intelligence} in the {Autonomous} {Navigation} of {Endovascular} {Interventions}: {A} {Systematic} {Review}},
	copyright = {Creative Commons Attribution 4.0 International},
	shorttitle = {Artificial {Intelligence} in the {Autonomous} {Navigation} of {Endovascular} {Interventions}},
	url = {https://arxiv.org/abs/2405.03305},
	doi = {10.48550/ARXIV.2405.03305},
	urldate = {2026-05-31},
	author = {Robertshaw, Harry and Karstensen, Lennart and Jackson, Benjamin and Sadati, Hadi and Rhode, Kawal and Ourselin, Sebastien and Granados, Alejandro and Booth, Thomas C},
	year = {2024},
}

@article{hu_comparison_1998,
	title = {Comparison of {Population}-{Averaged} and {Subject}-{Specific} {Approaches} for {Analyzing} {Repeated} {Binary} {Outcomes}},
	volume = {147},
	issn = {0002-9262, 1476-6256},
	url = {https://academic.oup.com/aje/article-lookup/doi/10.1093/oxfordjournals.aje.a009511},
	doi = {10.1093/oxfordjournals.aje.a009511},
	language = {en},
	number = {7},
	urldate = {2026-05-07},
	journal = {American Journal of Epidemiology},
	author = {Hu, F. B. and Goldberg, J. and Hedeker, D. and Flay, B. R. and Pentz, M. A.},
	month = apr,
	year = {1998},
	pages = {694--703},
}

@article{hagedorn_rural_2025,
	title = {Rural and {Urban} {Disparities} in {Access} to {Endovascular} {Thrombectomy} for {Large} {Vessel} {Occlusions} in {Colorado}},
	volume = {56},
	issn = {0039-2499, 1524-4628},
	url = {https://www.ahajournals.org/doi/10.1161/STROKEAHA.125.051542},
	doi = {10.1161/STROKEAHA.125.051542},
	language = {en},
	number = {11},
	urldate = {2026-01-04},
	journal = {Stroke},
	author = {Hagedorn, Danielle and Whaley, Michelle and Zuelke, Kristen and Ravare, Brandy and Jeppson, Kerri and Poisson, Sharon and Jones, William J. and O’Neil, Jonathan and Miller, Janice and Stringer, Donna and Pope, Brandon and Williams, Rachel and Eklund, Kelsey and Benson-Gallanis, N. Mary and Case, David and Farcas, Andra and Westensee, Nicole and Alexander, Mihaela and Reynolds, Wesley and McMahon, Jennifer and Dienst, Elizabeth and Scott, Jonathan W. and Ortiona, Melissa and Fanale, Chris and McDaneld, Logan and Johnson, Heather and Arnold, Kelly and Bartt, Russell and Jensen, Judd and Bennett, Alicia and Clausen, Suzanne and Bol, Michelle and Nieberlein, Amy and Dylla, Layne and Stein, Laura and Trivedi, Deepa and Shattuck, Jo and Sillau, Stefan H. and Scarbro, Sharon and Leppert, Michelle},
	month = nov,
	year = {2025},
	pages = {3209--3219},
}

@article{jovin_thrombectomy_2015,
	title = {Thrombectomy within 8 {Hours} after {Symptom} {Onset} in {Ischemic} {Stroke}},
	volume = {372},
	issn = {0028-4793, 1533-4406},
	url = {http://www.nejm.org/doi/10.1056/NEJMoa1503780},
	doi = {10.1056/NEJMoa1503780},
	language = {en},
	number = {24},
	urldate = {2025-10-27},
	journal = {New England Journal of Medicine},
	author = {Jovin, Tudor G. and Chamorro, Angel and Cobo, Erik and De Miquel, María A. and Molina, Carlos A. and Rovira, Alex and San Román, Luis and Serena, Joaquín and Abilleira, Sonia and Ribó, Marc and Millán, Mònica and Urra, Xabier and Cardona, Pere and López-Cancio, Elena and Tomasello, Alejandro and Castaño, Carlos and Blasco, Jordi and Aja, Lucía and Dorado, Laura and Quesada, Helena and Rubiera, Marta and Hernandez-Pérez, María and Goyal, Mayank and Demchuk, Andrew M. and Von Kummer, Rüdiger and Gallofré, Miquel and Dávalos, Antoni},
	month = jun,
	year = {2015},
	pages = {2296--2306},
}

@article{kimura_minimization_2016,
	title = {Minimization of {Akaike}'s {Information} {Criterion} in {Linear} {Regression} {Analysis} via {Mixed} {Integer} {Nonlinear} {Program}},
	copyright = {arXiv.org perpetual, non-exclusive license},
	url = {https://arxiv.org/abs/1606.05030},
	doi = {10.48550/ARXIV.1606.05030},
	urldate = {2025-10-27},
	publisher = {arXiv},
	author = {Kimura, Keiji and Waki, Hayato},
	year = {2016},
	note = {Version Number: 2},
}

@article{bates_fitting_2015,
	title = {Fitting {Linear} {Mixed}-{Effects} {Models} {Using} \textbf{lme4}},
	volume = {67},
	issn = {1548-7660},
	url = {http://www.jstatsoft.org/v67/i01/},
	doi = {10.18637/jss.v067.i01},
	language = {en},
	number = {1},
	urldate = {2025-10-27},
	journal = {Journal of Statistical Software},
	author = {Bates, Douglas and Mächler, Martin and Bolker, Ben and Walker, Steve},
	year = {2015},
}

@article{bilgin_impact_2025,
	title = {Impact of carotid artery tortuosity on mechanical thrombectomy outcomes: {A} systematic review},
	issn = {1971-4009, 2385-1996},
	shorttitle = {Impact of carotid artery tortuosity on mechanical thrombectomy outcomes},
	url = {https://journals.sagepub.com/doi/10.1177/19714009251317499},
	doi = {10.1177/19714009251317499},
	language = {en},
	urldate = {2025-10-26},
	journal = {The Neuroradiology Journal},
	author = {Bilgin, Cem and Gupta, Rishabh and Orscelik, Atakan and Hassankhani, Amir and Senol, Yigit Can and Kobeissi, Hassan and Ghozy, Sherief and Kadirvel, Ramanathan and Brinjikji, Waleed and Kallmes, David F.},
	month = feb,
	year = {2025},
	pages = {19714009251317499},
}

@article{miki_combined_2021,
	title = {Combined {Technique} {Thrombectomy} with a {Long} {Balloon}-{Guiding} {Catheter} and {Long} {Sheath} {Aids} in {Rapid} and {Stable} {Recanalization} in {Patients} with {Anterior} {Circulation} {Acute} {Ischemic} {Stroke}},
	volume = {15},
	issn = {1882-4072, 2186-2494},
	url = {https://www.jstage.jst.go.jp/article/jnet/15/5/15_oa.2020-0041/_article},
	doi = {10.5797/jnet.oa.2020-0041},
	language = {en},
	number = {5},
	urldate = {2025-10-26},
	journal = {Journal of Neuroendovascular Therapy},
	author = {Miki, Kazunori and Aizawa, Yuki and Fujii, Shoko and Karakama, Jun and Fujita, Kyohei and Sasaki, Yoshiyuki and Nemoto, Shigeru and Sumita, Kazutaka},
	year = {2021},
	pages = {281--287},
}

@inproceedings{lahlouh_automated_2023,
	address = {Sydney, Australia},
	title = {Automated {Aortic} {Anatomy} {Analysis}: from {Image} to {Clinical} {Indicators}},
	copyright = {https://doi.org/10.15223/policy-029},
	isbn = {979-8-3503-2447-1},
	shorttitle = {Automated {Aortic} {Anatomy} {Analysis}},
	url = {https://ieeexplore.ieee.org/document/10340921/},
	doi = {10.1109/EMBC40787.2023.10340921},
	urldate = {2025-10-26},
	booktitle = {2023 45th {Annual} {International} {Conference} of the {IEEE} {Engineering} in {Medicine} \& {Biology} {Society} ({EMBC})},
	publisher = {IEEE},
	author = {Lahlouh, M. and Chenoune, Y. and Blanc, R. and Piotin, M. and Escalard, S. and Fahed, R. and Szewczyk, J. and Passat, N.},
	month = jul,
	year = {2023},
	pages = {1--5},
}

@article{bhogal_initial_2025,
	title = {The initial experience with the {Walrus} balloon guide catheter – {Results} from two high-volume thrombectomy centres},
	issn = {1591-0199, 2385-2011},
	url = {https://journals.sagepub.com/doi/10.1177/15910199251336935},
	doi = {10.1177/15910199251336935},
	language = {en},
	urldate = {2025-10-26},
	journal = {Interventional Neuroradiology},
	author = {Bhogal, Pervinder and Dhillon, Permesh Singh and Flood, Richard and Lewis, Martin and Podlasek, Anna and Wong, Ken and Lansley, Joseph and Makalanda, Levansri and Minks, David and Spooner, Oliver and Mortimer, Alex},
	month = may,
	year = {2025},
	pages = {15910199251336935},
}

@article{ando_factors_2025,
	title = {Factors of {Difficult} {Guiding} {Catheter} {Access} in {Mechanical} {Thrombectomy} for {Acute} {Ischemic} {Stroke} in the {Anterior} {Circulation}},
	volume = {19},
	issn = {1882-4072, 2186-2494},
	url = {https://www.jstage.jst.go.jp/article/jnet/19/1/19_oa.2024-0108/_article},
	doi = {10.5797/jnet.oa.2024-0108},
	language = {en},
	number = {1},
	urldate = {2025-10-26},
	journal = {Journal of Neuroendovascular Therapy},
	author = {Ando, Kazuhiro and Kikuchi, Bumpei and Watanabe, Jun and Takino, Toru and Mouri, Yoshihiro and Watabe, Yuki and Shida, Kazuki and Yamashita, Shinya},
	year = {2025},
	pages = {n/a},
}

@article{snelling_unfavorable_2018,
	title = {Unfavorable {Vascular} {Anatomy} {Is} {Associated} with {Increased} {Revascularization} {Time} and {Worse} {Outcome} in {Anterior} {Circulation} {Thrombectomy}},
	volume = {120},
	issn = {18788750},
	url = {https://linkinghub.elsevier.com/retrieve/pii/S1878875018319934},
	doi = {10.1016/j.wneu.2018.08.207},
	language = {en},
	urldate = {2025-10-26},
	journal = {World Neurosurgery},
	author = {Snelling, Brian M. and Sur, Samir and Shah, Sumedh S. and Chen, Stephanie and Menaker, Simon A. and McCarthy, David J. and Yavagal, Dileep R. and Peterson, Eric C. and Starke, Robert M.},
	month = dec,
	year = {2018},
	pages = {e976--e983},
}

@article{moosa_benchmarking_2025,
	title = {Benchmarking reinforcement learning algorithms for autonomous mechanical thrombectomy},
	volume = {20},
	issn = {1861-6429},
	url = {https://link.springer.com/10.1007/s11548-025-03360-x},
	doi = {10.1007/s11548-025-03360-x},
	language = {en},
	number = {6},
	urldate = {2025-10-22},
	journal = {International Journal of Computer Assisted Radiology and Surgery},
	author = {Moosa, Farhana and Robertshaw, Harry and Karstensen, Lennart and Booth, Thomas C. and Granados, Alejandro},
	month = apr,
	year = {2025},
	pages = {1231--1238},
}

@article{robertshaw_autonomous_2024,
	title = {Autonomous navigation of catheters and guidewires in mechanical thrombectomy using inverse reinforcement learning},
	volume = {19},
	issn = {1861-6429},
	url = {https://link.springer.com/10.1007/s11548-024-03208-w},
	doi = {10.1007/s11548-024-03208-w},
	language = {en},
	number = {8},
	urldate = {2025-10-22},
	journal = {International Journal of Computer Assisted Radiology and Surgery},
	author = {Robertshaw, Harry and Karstensen, Lennart and Jackson, Benjamin and Granados, Alejandro and Booth, Thomas C.},
	month = jun,
	year = {2024},
	pages = {1569--1578},
}

@article{robertshaw_reinforcement_2025,
	title = {Reinforcement learning for safe autonomous two-device navigation of cerebral vessels in mechanical thrombectomy},
	volume = {20},
	issn = {1861-6429},
	url = {https://link.springer.com/10.1007/s11548-025-03339-8},
	doi = {10.1007/s11548-025-03339-8},
	language = {en},
	number = {6},
	urldate = {2025-10-22},
	journal = {International Journal of Computer Assisted Radiology and Surgery},
	author = {Robertshaw, Harry and Jackson, Benjamin and Wang, Jiaheng and Sadati, Hadi and Karstensen, Lennart and Granados, Alejandro and Booth, Thomas C.},
	month = apr,
	year = {2025},
	pages = {1077--1086},
}

@article{feigin_world_2025,
	title = {World {Stroke} {Organization}: {Global} {Stroke} {Fact} {Sheet} 2025},
	volume = {20},
	issn = {1747-4930, 1747-4949},
	shorttitle = {World {Stroke} {Organization}},
	url = {https://journals.sagepub.com/doi/10.1177/17474930241308142},
	doi = {10.1177/17474930241308142},
	language = {en},
	number = {2},
	urldate = {2025-10-16},
	journal = {International Journal of Stroke},
	author = {Feigin, Valery L and Brainin, Michael and Norrving, Bo and Martins, Sheila O and Pandian, Jeyaraj and Lindsay, Patrice and F Grupper, Maria and Rautalin, Ilari},
	month = feb,
	year = {2025},
	pages = {132--144},
}

\end{document}